\documentclass[letterpaper, 10 pt, conference]{ieeeconf}  % Comment this line out if you need a4paper

\IEEEoverridecommandlockouts                              % This command is only needed if
\usepackage{array}
\usepackage{booktabs}
\usepackage{censor}
\usepackage[bottom]{footmisc}
\usepackage{gensymb}
\usepackage{graphicx}
\usepackage{hyperref}
\usepackage{lipsum}
\usepackage{multirow}
\usepackage{stfloats}
\usepackage{subcaption}
\usepackage{textcomp}
\usepackage{url}
\usepackage{xcolor}
\usepackage{cleveref}  % must be loaded after hyperref

\crefname{figure}{Fig.}{Figs.}
\Crefname{figure}{Fig.}{Figs.}
\crefname{table}{Tab.}{Tabs.}
\Crefname{table}{Tab.}{Tabs.}
\crefname{section}{Sec.}{Secs.}
\Crefname{section}{Sec.}{Secs.}
\crefname{subsection}{Sec.}{Secs.}
\Crefname{subsection}{Sec.}{Secs.}
\crefname{appendix}{App.}{Apps.}
\Crefname{appendix}{App.}{Apps.}

\newcommand{\orcid}[1]{\raisebox{1pt}{\href{https://orcid.org/#1}{\includegraphics[height=10pt]{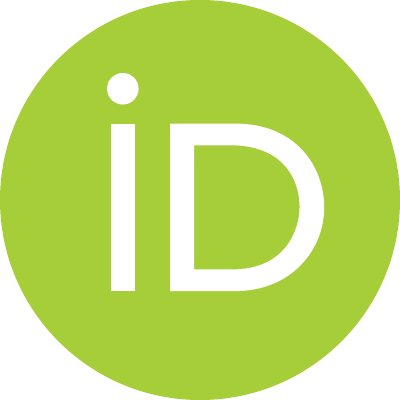}}}}

\title{\LARGE \bf
    RACAS: Controlling Diverse Robots
    With a Single Agentic System
}

\author{
    Dylan R.~Ashley,\hspace{-0.025em}$^{*}$\hspace{0.05em}$^{1,2,3,4}$
    Jan Przepi\'ora,\hspace{-0.025em}$^{*}$\hspace{0.05em}$^{5,6}$
    Yimeng Chen,\hspace{-0.025em}$^{*}$\hspace{0.05em}$^{1}$\\
    Ali Abualsaud,\hspace{0.05em}$^{5}$
    Nurzhan Yesmagambet,\hspace{0.05em}$^{5}$
    Shinkyu Park,$^{5}$
    Eric Feron,$^{5}$ and
    J\"{u}rgen Schmidhuber\hspace{0.2em}$^{1,2,3,4}$
    \thanks{$^{*}$Equal contribution.
    $^{1}$Center of Excellence in Generative AI, King Abdullah University of Science and Technology (KAUST), Saudi Arabia.
    $^{2}$Dalle Molle Institute for Artificial Intelligence Research (IDSIA), Switzerland.
    $^{3}$Universit\`{a} della Svizzera italiana (USI), Switzerland.
    $^{4}$Scuola universitaria professionale della Svizzera italiana (SUPSI), Switzerland.
    $^{5}$Robotics, Intelligent Systems, and Control Lab, King Abdullah University of Science and Technology (KAUST), Saudi Arabia.
    $^{6}$Department of Process Control, AGH University of Krakow, Poland.
    Correspondence to \href{mailto:dylan.ashley@usi.ch}{\tt dylan.ashley@usi.ch}, \href{mailto:jan.przepiora@kaust.edu.sa}{\tt jan.przepiora@kaust.edu.sa}, or \href{mailto:yimeng.chen@kaust.edu.sa}{\tt yimeng.chen@kaust.edu.sa}.}
}

\begin{document}

\bstctlcite{IEEEexample:BSTcontrol}

\maketitle
\thispagestyle{empty}
\pagestyle{empty}

%\setcounter{footnote}{6}.  % so that footnotes start after the last affiliation footnote

%%%%%%%%%%%%%%%%%%%%%%%%%%%%%%%%%%%%%%%%%%%%%%%%%%%%%%%%%%%%%%%%%%%%%%%%%%%%%%%%

\begin{figure*}[b!]  % bottom of the first page or top of the second page
    \centering
    \includegraphics[width=.83\textwidth]{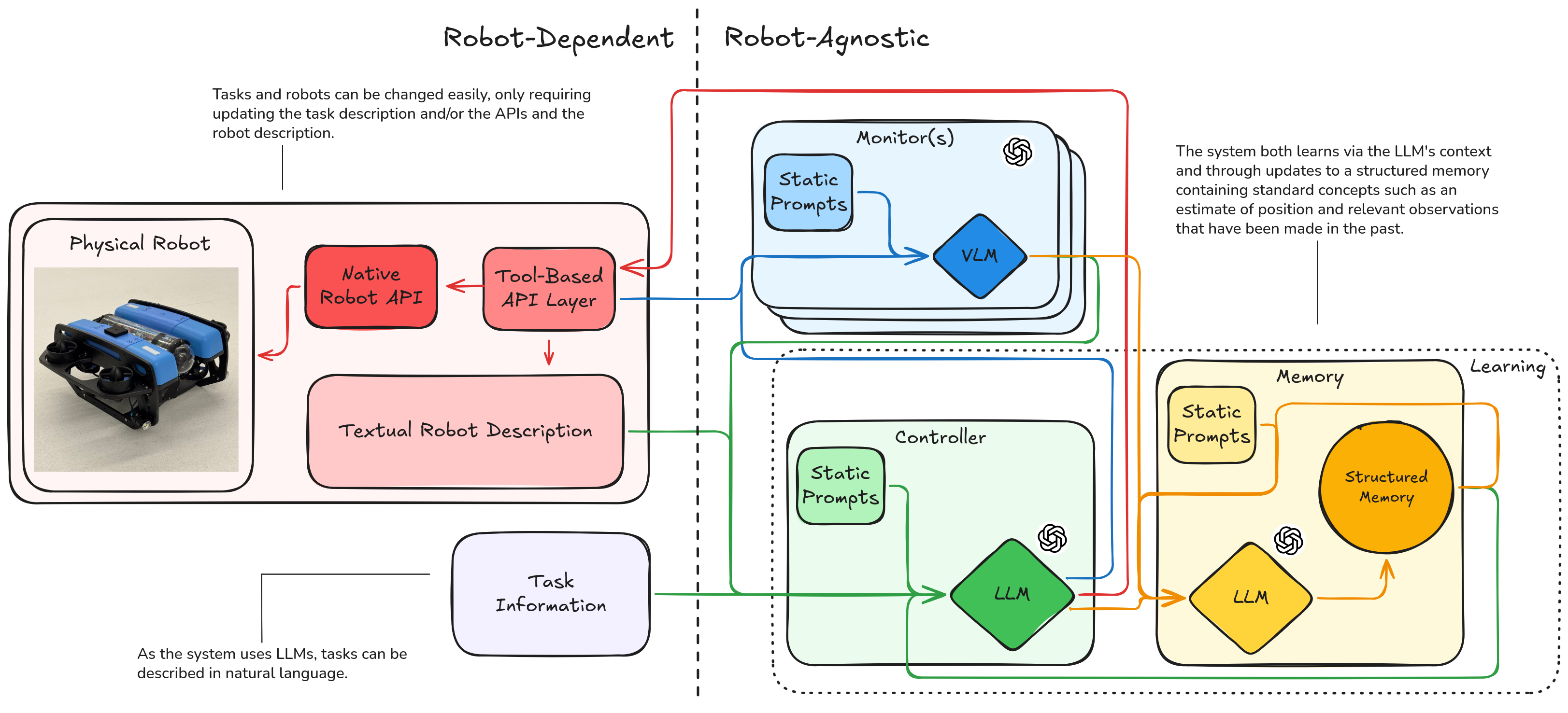}
    \caption{The basic design of the system tries to minimize the extent of changes needed to apply it to different robots. The majority of the complexity of adapting to different robot platforms is offloaded to large AI models augmented with a dynamic structured memory system.}
    \label{fig:system_diagram}  % to be referenced in methodology section
\end{figure*}

\begin{abstract}
    % Deploying autonomous robots across diverse platforms typically requires embodiment-specific training data, hand-engineered state representations, or platform-dependent control modules, limiting generalization to structurally similar robots. We introduce RACAS (Robot-Agnostic Control via Agentic System), a single agentic system that controls heterogeneous robot embodiments without retraining or code modification. RACAS decomposes closed-loop control into three cooperative modules: an LLM-based Controller that selects actions through reasoning, a set of Monitors that provide task-adaptive visual perception via query-driven VLM inference, and a Memory Curator that maintains a structured, bounded representation of accumulated experience through incremental rewriting. All modules communicate exclusively through natural language, and all embodiment-specific knowledge is encapsulated in declarative prompt configurations rather than in learned parameters. We evaluate RACAS on three platforms that differ radically in morphology, maneuverability, and operating environment: a novel robotic limb, a wheeled ground robot, and an underwater vehicle.
    Many robotic platforms expose an API through which external software can command their actuators and read their sensors.
    However, transitioning from these low-level interfaces to high-level autonomous behaviour requires a complicated pipeline, whose components demand distinct areas of expertise.
    Existing approaches to bridging this gap either require retraining for every new embodiment or have only been validated across structurally similar platforms.
    We introduce RACAS (Robot-Agnostic Control via Agentic Systems), a cooperative agentic architecture in which three LLM/VLM-based modules (Monitors, a Controller, and a Memory Curator) communicate exclusively through natural language to provide closed-loop robot control.
    RACAS requires only a natural language description of the robot, a definition of available actions, and a task specification; no source code, model weights, or reward functions need to be modified to move between platforms.
    We evaluate RACAS on several tasks using a wheeled ground robot, a recently published novel multi-jointed robotic limb, and an underwater vehicle.
    RACAS consistently solved all assigned tasks across these radically different platforms, demonstrating the potential of agentic AI to substantially reduce the barrier to prototyping robotic solutions.
    % As the underlying foundation models continue to improve, we expect this approach to become increasingly viable for a wider range of tasks and platforms.
\end{abstract}

%%%%%%%%%%%%%%%%%%%%%%%%%%%%%%%%%%%%%%%%%%%%%%%%%%%%%%%%%%%%%%%%%%%%%%%%%%%%%%%%

\section{INTRODUCTION}\label{sec:introduction}

% ~ 1 page

\begin{figure*}[t!]
    \centering
    \vspace{4pt}
    \subcaptionbox{Clearpath Dingo}{\includegraphics[height=1.45in]{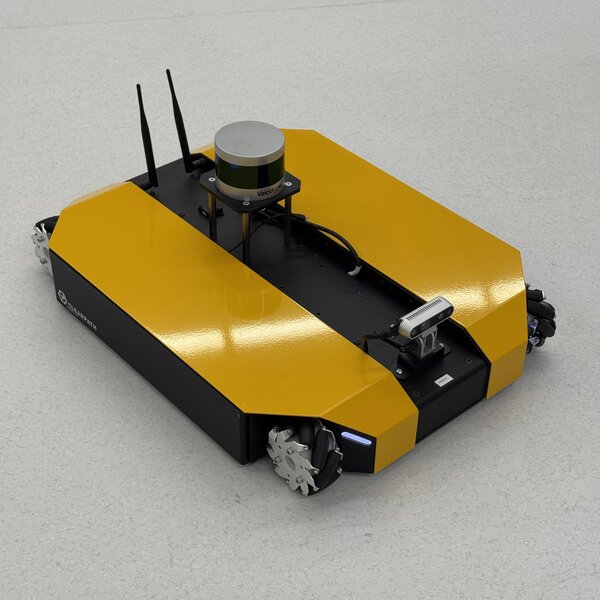}}
    \hspace{.02\textwidth}
    \subcaptionbox{Alhakami et al. Limb}{\includegraphics[height=1.45in]{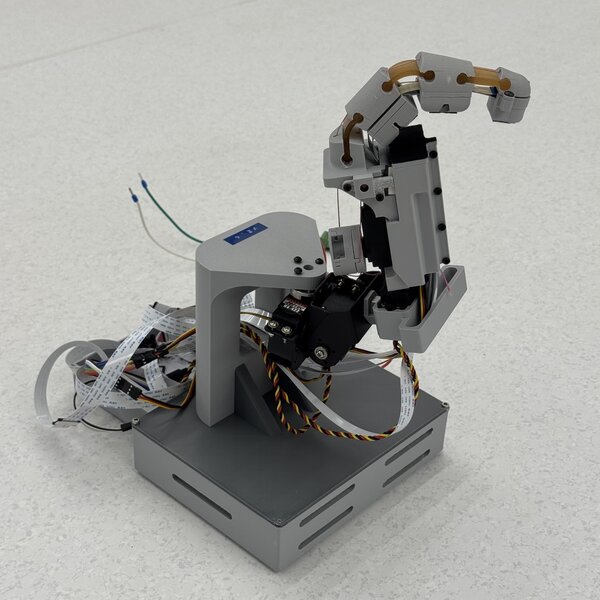}}
    \hspace{.02\textwidth}
    \subcaptionbox{BlueRobotics BlueROV2}{\includegraphics[height=1.45in]{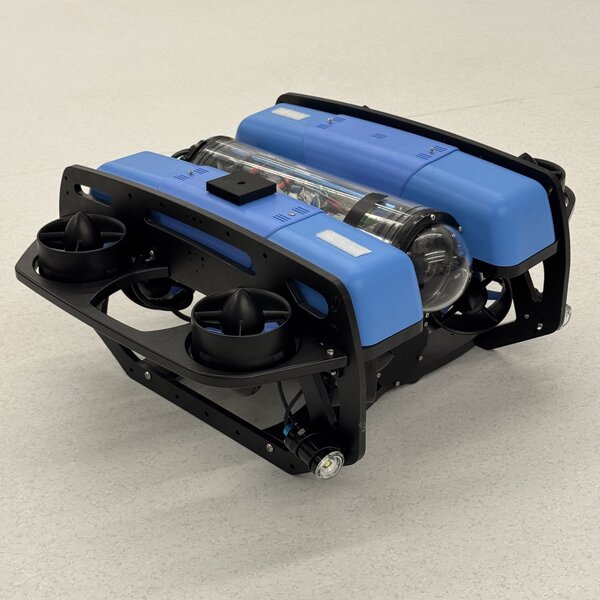}}
    \caption{The three real-world robotic platforms the system from \cref{fig:system_diagram} was applied to.}
    \label{fig:robots}
\end{figure*}

Most robots deployed in operational settings today come with a programmatic API through which external software can command their actuators and read their sensors.
However, applying a robot to a given problem requires a complicated pipeline that moves from these relatively low-level control commands to high-level behavior.
This pipeline generally contains elements such as perception modules that extract task-relevant state from raw sensor streams, planning algorithms that reason over that state, and control policies that map this reasoning to platform-specific motor commands.
Importantly, the expertise needed to develop the low-level and high-level parts of this pipeline comes from disjoint areas of specialization; hardware and firmware engineers who understand the robot's kinematics, sensor characteristics, and communication protocols rarely specialize in the planning and learning algorithms that govern task execution, and vice versa.
This gap creates a barrier for applying robots to various tasks that scales poorly with the growing diversity of commercially available platforms.

Existing approaches to bridging the gap between low-level motor commands and high-level control algorithms fall into two broad categories.
The first approach is to train end-to-end policies on robot-specific demonstration data.
While effective within the training distribution, these methods require data collection and retraining for every new embodiment and degrade when the target platform differs structurally from the training fleet~\cite{brohan2022rt1,brohan2023rt2,bousmalis2023robocat}.
The second approach utilizes large language models (LLM) or vision language models (VLM) reasoning around robot API libraries, allowing natural language to serve as a common interface between high-level intent and low-level commands~\cite{vemprala2024chatgptforrobotics,ahn2022saycan,liang2023codeaspolicies}.
These frameworks reduce embodiment-specific engineering, but have so far been validated only on structurally highly-similar platforms (e.g., different manipulator arms) in controlled laboratory settings.
To the best of our knowledge, prior systems demonstrated zero-training generalization across radically heterogeneous platforms where morphology, dynamics, and operating environments differ fundamentally.

While a custom-engineered system would inevitably be needed for production-grade deployment (if only because of the stochastic nature of the alternative), we investigate whether a general agentic AI-based system can provide robot-agnostic closed-loop control using only the programmatic interfaces that robots already expose.
To this end, we introduce RACAS (Robot-Agnostic Control via Agentic Systems), a relatively simple agentic architecture that decomposes control into three cooperative LLM/VLM-based modules that communicate exclusively through natural language: a controller, a set of monitors, and a memory curator (\cref{fig:system_diagram}).
This system requires only a natural language description of the robot, a structured definition of available actions (that is to say, a light abstraction layer over the API and its documentation converting commands like \texttt{servo\_drive(1m)} to \texttt{move\_forwards}), and a description of the task and what considerations the robot should have when executing it.
No source code, model weights, or reward functions are modified; all embodiment-specific and task-specific knowledge is confined to these declarative prompt configurations.
Learning occurs through the memory curator, allowing the controller to learn to query the monitor (in the style of Learning to Think~\cite{schmidhuber2015learning} or NLSOMs~\cite{zhuge2023mindstorms,zhuge2025mindstorms}).

We evaluate RACAS on platforms that differ radically in morphology, movement capabilities, and operating environment (\cref{fig:robots}): a wheeled ground robot (Clearpath Dingo\footnote{\url{https://clearpathrobotics.com/dingo-indoor-mobile-robot/}}) in both simulation and in the real world, a modified recently-published novel robotic limb by Alhakami et al.~\cite{alhakami2025towards} (which an LLM would thus have no knowledge of), and an underwater remotely operated vehicle (BlueRobotics BlueROV2\footnote{\url{https://bluerobotics.com/store/rov/bluerov2/}}).
In all cases, the unmodified RACAS successfully completes the assigned tasks.
Altogether, this demonstrates the potential of agentic AI to substantially reduce the barrier to prototyping and testing robotic solutions across diverse platforms.

\textbf{Our main contributions are as follows:}
\begin{enumerate}
    \item We propose RACAS, a cooperative multi-module agentic architecture for closed-loop robot control in which all communication occurs in natural language.
    \item We show that all embodiment-specific and task-specific knowledge can be confined to declarative prompt configurations, requiring no modifications to source code, model weights, or reward functions to adapt the system to a new platform.
    \item We demonstrate---to the best of our knowledge---the first zero-training generalization of a single control framework across three radically heterogeneous robot platforms, including a very recent robotic platform that an LLM would have no prior knowledge of.
\end{enumerate}

The source code for all our experiments (including prompts) is available at \url{https://github.com/janprz11/robot-agnostic-control}

\section{RELATED WORK}\label{sec:background_and_related_work}
% ~ 0.5 pages

\subsection{LLM/VLM-Based Robotic Control}

Recent work has increasingly leveraged large language models (LLMs) and vision-language models (VLMs) as core components of robotic control pipelines. A foundational paradigm is to use the LLM as a high-level planner whose outputs are grounded through low-level skill libraries or affordance models. SayCan~\cite{ahn2022saycan} pairs a pre-trained LLM with a learned value function that scores the feasibility of candidate actions, enabling a mobile manipulator to execute long-horizon household instructions from a fixed skill set. Two concurrent lines of work further develop this paradigm along complementary axes. Code-as-Policies~\cite{liang2023codeaspolicies} has the LLM generate executable Python code that directly calls robot APIs, improving interpretability and enabling reactive feedback loops within the generated programs, though it does not support iterative LLM re-planning when the generated code fails. Inner Monologue~\cite{huang2022innermonologue} instead incorporates perceptual feedback, such as success detection and scene descriptions, into the LLM's planning loop, allowing dynamic re-planning upon failure. More recent frameworks, such as RobotIQ~\cite{raptis2025robotiq} and ChatGPT for Robotics~\cite{vemprala2024chatgptforrobotics}, demonstrate modular integration of LLMs with robot control stacks across multiple robot types. While both integrate dedicated visual perception modules, their reasoning remains centralized in a single text-based LLM rather than leveraging vision-language models as core reasoning components.
% (e.g., AR marker tracking and YOLOv8-based object detection) via API libraries

To overcome the limitations of monolithic architectures, several works explore multi-agent designs. MALMM~\cite{singh2024malmm} decomposes robot control into three specialized LLM agents, including Planner, Coder, and Supervisor, which collaboratively generate and verify manipulation plans in a zero-shot setting. RoCo~\cite{mandi2024roco} employs an LLM-mediated dialogue to coordinate multiple robots, where agents negotiate task allocation through natural language exchange. LLM2Swarm~\cite{strobel2024llm2swarm} scales this concept to robot swarms by equipping each agent with a local LLM instance for real-time linguistic coordination and emergent collective behavior. On the perception side, Manipulate-Anything~\cite{duan2024manipulateanything} constructs a multi-stage VLM pipeline that identifies objects, generates sub-tasks, executes actions, and verifies outcomes through vision, demonstrating the value of tight visual grounding. VADER~\cite{ahn2024vader} similarly leverages visual question answering to detect execution failures and dynamically delegate sub-tasks to other robots or humans. Despite this progress, existing multi-agent systems either consist entirely of text-only LLMs with limited visual understanding~\cite{singh2024malmm,mandi2024roco} or employ VLMs only at specific pipeline stages within a single-domain setting~\cite{duan2024manipulateanything,ahn2024vader}. In contrast, our work focuses on a general-purpose architecture in which a VLM and an LLM operate as functionally complementary agents.

\subsection{Cross-Embodiment Generalization}
Generalization across diverse embodiments without task-specific retraining is a central aspiration of robotics.
Model-Agnostic Meta-Learning (MAML)~\cite{finn2017maml}, for example, attempts to achieve this by optimizing initial model parameters so that a small number of gradient steps suffice to adapt to a new task.
Despite improving sample efficiency at test time, early meta-reinforcement learning methods still require extensive meta-training across curated task distributions and struggle to generalize across semantically diverse tasks~\cite{yu2020metaworld,beck2025tutorial}.
Critically, they do not naturally extend to cross-embodiment settings without explicit hardware encodings or additional data collection for each new morphology.
With the advent of foundation models, some have suggested such models might be the key to achieving this~\cite{firoozi2025foundation}.
% , and has been applied to tasks like locomotion. Subsequent extensions enable real-world online dynamics adaptation in sub-second time scales~\cite{nagabandi2019learning}, off-policy meta-reinforcement learning with probabilistic task inference~\cite{rakelly2019pearl}, and cross-robot transfer via hardware-conditioned policies~\cite{chen2018hardware}.
Data-driven approaches have pursued this through scale: RT-1~\cite{brohan2022rt1} trains a Transformer policy on over 130k demonstrations from a fleet of identical robots, achieving 97\% success on seen tasks, but degrading on novel instructions outside the training distribution. RT-2~\cite{brohan2023rt2} improves generalization by co-fine-tuning a vision-language model on both internet-scale image-text data and robot demonstrations, enabling emergent semantic reasoning (e.g., picking up an object that could be used as an improvised hammer), yet still requires robot-specific action data for training. RoboCat~\cite{bousmalis2023robocat} demonstrates cross-embodiment transfer across different robotic arms by iteratively fine-tuning and self-generating new training data, but each new embodiment demands additional data collection and retraining cycles. At a larger scale, PaLM-E~\cite{driess2023palme} and Gato~\cite{reed2022gatox} show that a single pre-trained large model can handle diverse modalities and tasks. Framework-level approaches such as ChatGPT for Robotics~\cite{vemprala2024chatgptforrobotics} sidestep embodiment-specific training by wrapping LLM reasoning around robot API libraries through a common language interface. However, none of them has been validated beyond controlled laboratory settings with standard robot platforms.
% However, they only validated on conventional robot platforms in structured environments.
% PaLM-E through a 562B-parameter multimodal architecture whose computational demands preclude real-time on-robot deployment, and Gato through a unified Transformer trained on 604 tasks that, while demonstrating broad competence, underperforms task-specific models on individual benchmarks.

A notable gap remains: prior approaches typically require training data or fine-tuning specific to the target embodiment, or have only been validated across structurally similar platforms (e.g., different manipulator arms). Demonstration of zero-training generalization across radically heterogeneous platforms remains rare, where morphology, dynamics, and operating environments differ fundamentally.

\subsection{Memory and Adaptation Mechanisms}

Effective agentic behavior requires not only planning but also the ability to accumulate and retrieve experiential knowledge over time. Several works have explored structured knowledge representations to support LLM-based robot planning, though each addresses only a narrow aspect of the information a robot needs to retain. SayPlan~\cite{rana2023sayplan} augments an LLM with a 3D scene graph encoding spatial relationships among rooms and objects; while the framework performs iterative scene graph search and re-grounding during planning, the graph itself is externally constructed and does not encode the robot's own capabilities or experiential knowledge. LLM-GROP~\cite{ding2023tampobjectrearrangement} and ProgPrompt~\cite{singh2023progprompt} similarly encode environment and action specifications in symbolic or programmatic form, yet these are manually defined per domain and do not evolve with experience.

Other works construct representations that are ephemeral or unstructured. VoxPoser~\cite{huang2023voxposer} composes 3D voxel-based value maps through LLM-generated code that queries a VLM for spatial grounding, but these maps are discarded after each task. Inner Monologue~\cite{huang2022innermonologue} maintains a running log of actions and perceptual feedback within an episode, enabling reactive re-planning, yet this trace does not persist across tasks and carries no mechanism for long-term knowledge accumulation.
RONAR~\cite{wang2024ronar} uses an LLM to narrate the robot's sensory experience as natural language traces, improving human situational awareness, but these narratives serve a descriptive role and do not feed back into the control loop for future decision-making. Similarly, MOO~\cite{stone2023moo} leverages a frozen VLM to extract object localization masks for open-vocabulary manipulation and conditions a Transformer-based policy on observation history, but this implicit temporal context carries no structured or persistent memory.
A separate line of work explores self-reflection as an adaptation mechanism. BUMBLE~\cite{shah2024bumble} combines short-term memory for online re-planning with a long-term memory of human-annotated failure cases. Most recently, PragmaBot~\cite{qu2026pragmabot} uses a VLM for both planning and visual success verification, enabling fully autonomous self-reflection.
A limitation shared by all these approaches is their lack of an organized, persistent memory that captures the distinct types of knowledge a robot accumulates in operation.
% In practice, a robot operating in the real world must simultaneously maintain knowledge about its own embodiment (e.g., kinematic constraints, actuation limits, learned behavioral patterns), the environment (e.g., spatial layout, object properties, task-relevant context), and its interaction history (e.g., past action trajectories, their outcomes, and failure modes). No existing work structures test-time experience along these complementary dimensions, nor uses such structured traces to adapt behavior to new embodiments and environments without retraining.

\begin{figure*}[ht]
    \centering
    \vspace{4pt}
    \includegraphics[width=0.88\linewidth]{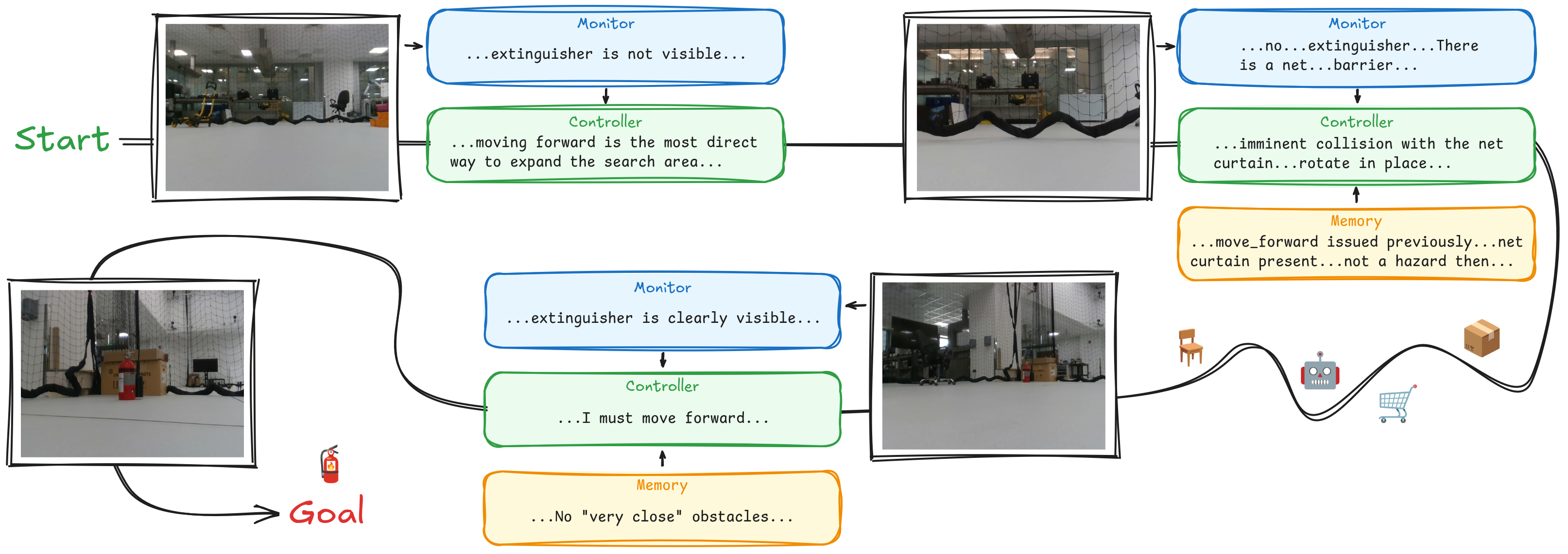}
    \caption{An example of RACAS's reasoning
    process during one of the Dingo (real-world) runs.}
    \label{fig:example_diagram}
\end{figure*}

\section{METHODOLOGY}\label{sec:methodology}
% ~ 1.5 pages

We present a framework for language-driven robotic control that operates across heterogeneous robot embodiments without retraining or code modification. The core idea is to decompose closed-loop control into three cooperative LLM pipelines: perception, decision, and memory, connected through a unified natural language interface, and to encapsulate all embodiment-specific knowledge in declarative prompt configurations rather than in learned parameters or engineered modules.

\subsection{Problem Formulation}
\label{subsec:problem}

We consider the problem of controlling a robot whose morphology is known only through a natural language description $\mathcal{D}$, a text-defined set of admissible actions $\mathcal{A} = \{a_1, \dots, a_K\}$, and a stream of raw images from one or more onboard cameras $\{I_1^{(c)}, \dots, I_t^{(c)}\}_{c=1}^{C}$, to achieve a user-specified task $\tau$. No dynamics model, kinematic parameters, or embodiment-specific learned weights are provided.

At each time step $t$, the system must select a single action $a_t \in \mathcal{A}$ based on the current visual observations, the task specification, and all prior interaction history. Unlike conventional formulations where the state space $\mathcal{S}$ and action space $\mathcal{A}$ are fixed-dimensional vectors tied to a specific robot, here both are represented symbolically in natural language and vary across embodiments. The action space for a multi-joint manipulator may contain eight primitives, while an underwater vehicle may expose six. The framework accommodates this variation without architectural change.

\subsection{System Architecture}
\label{subsec:architecture}

The system consists of three LLM-based modules: a \emph{Controller}, a set of \emph{Monitors}, and a \emph{Memory Curator}. They are orchestrated in a closed loop around a hardware abstraction layer (Fig.~\ref{fig:system_diagram}). At each step $t$, the loop proceeds as follows:

\begin{enumerate}
    \item The Controller generates a targeted visual query $q_t$ based on the current task state.
    \item Each Monitor $m_c$ processes its camera image $I_t^{(c)}$ conditioned on $q_t$ and returns a natural language scene description $o_t^{(c)}$.
    \item The Controller receives all monitor observations $\{o_t^{(c)}\}$, reasons over them, and selects an action $a_t$.
    \item The action is dispatched to the robot through the hardware abstraction layer.
    \item The Memory Curator integrates the interaction record into a persistent environment memory $\mathcal{M}_t$.
\end{enumerate}

All three modules communicate exclusively through natural language. Adapting the system to a new robot requires only replacing the prompt configuration files (the robot description $\mathcal{D}$, the action definitions $\mathcal{A}$, and the environment context), with no modification to the control logic itself.

The Controller is re-initialized at each step with a dynamically composed system prompt assembled from six components: the robot description $\mathcal{D}$, the action interface $\mathcal{A}$, the environment memory $\mathcal{M}_{t-1}$ accumulated from prior steps (Sec.~\ref{subsec:memory}), the proprioceptive state $\mathbf{s}_t$ (joint displacements maintained by an internal position tracker), the action history $H_t$, and the task specification $\tau$. In step~1, the Controller reasons over this context to produce the visual query $q_t$; in step~3, it receives the monitor observations and outputs a reasoning trace followed by a single action $a_t$. \cref{fig:example_diagram} showcases the reasoning process with RACAS during a real run of the Dingo experiments.

\begin{figure*}[t]
    \centering
    \vspace{4pt}
    \subcaptionbox{Alhakami et al. Limb setup\label{fig:limb_setup}}{\includegraphics[height=1.65in]{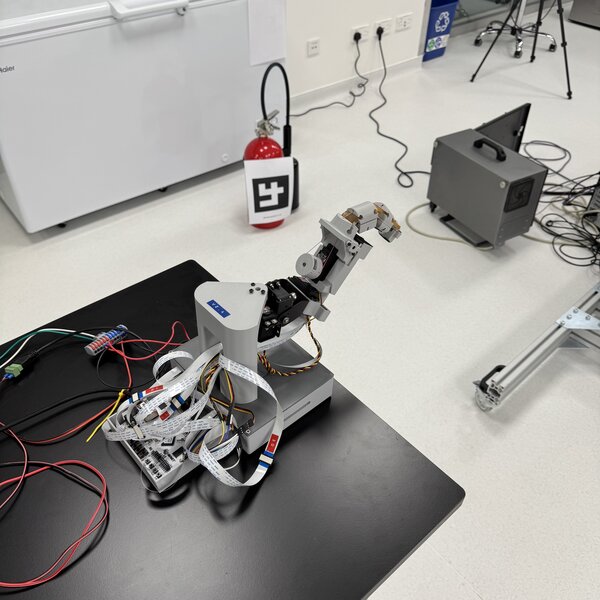}}
    \subcaptionbox{Dingo (simulation) setup\label{fig:simulation_setup}}{\includegraphics[height=1.65in]{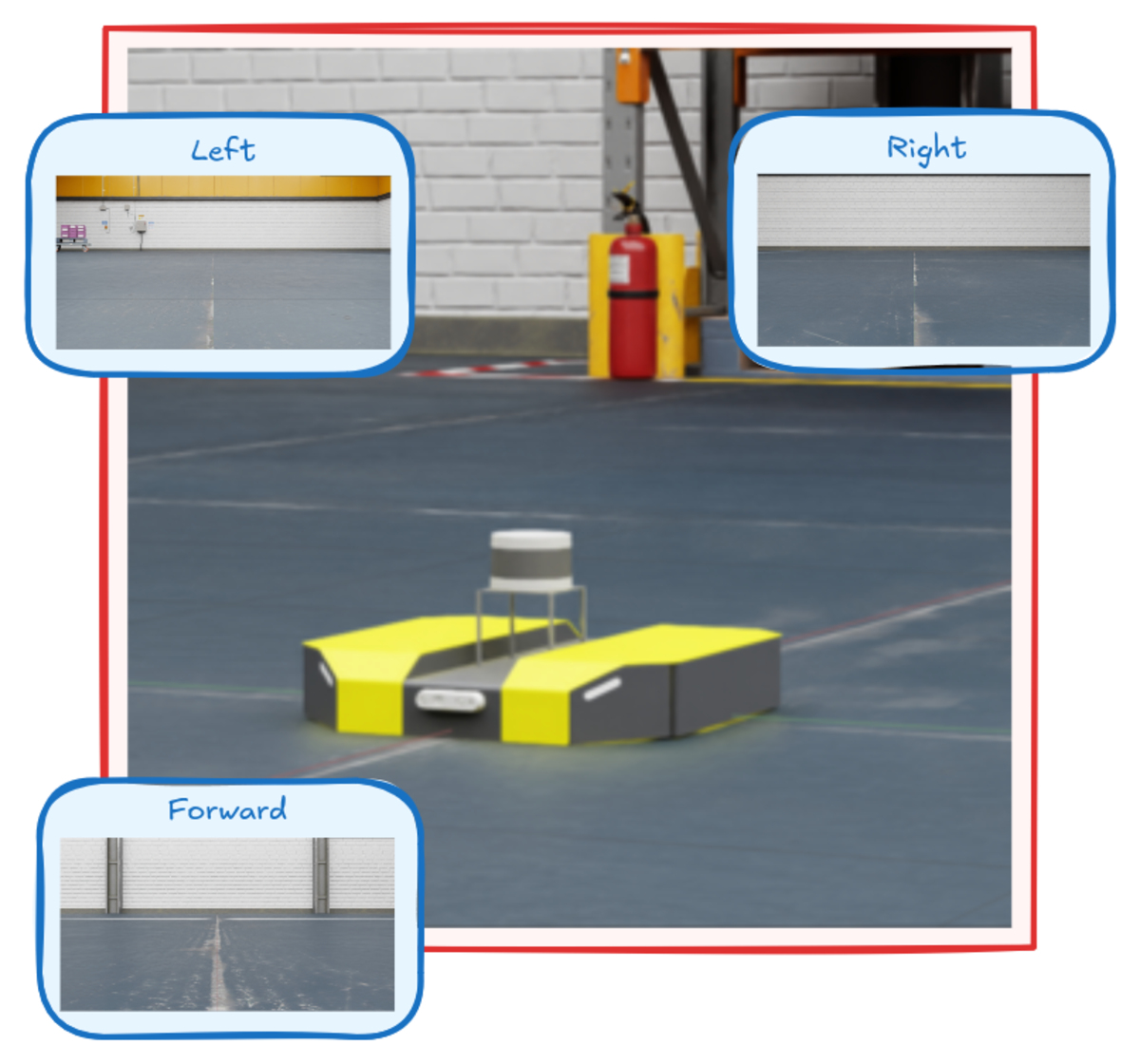}}
    \subcaptionbox{Dingo (real-world) setup\label{fig:real_world_setup}}{\includegraphics[height=1.65in]{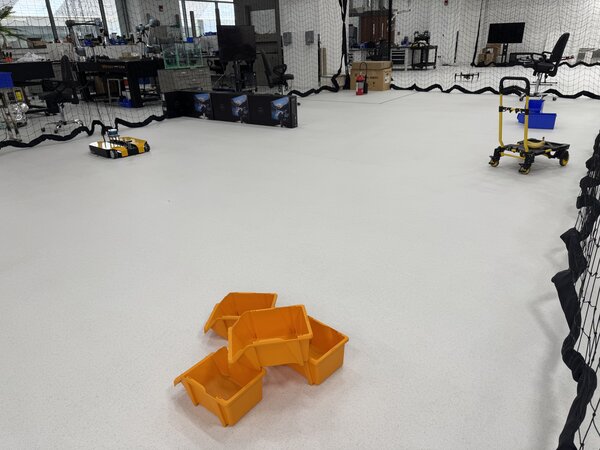}}
    \\\vspace{0.5em}
    \subcaptionbox{BlueROV2 (small tank) setup\label{fig:small_tank_setup}}{\includegraphics[height=1.65in]{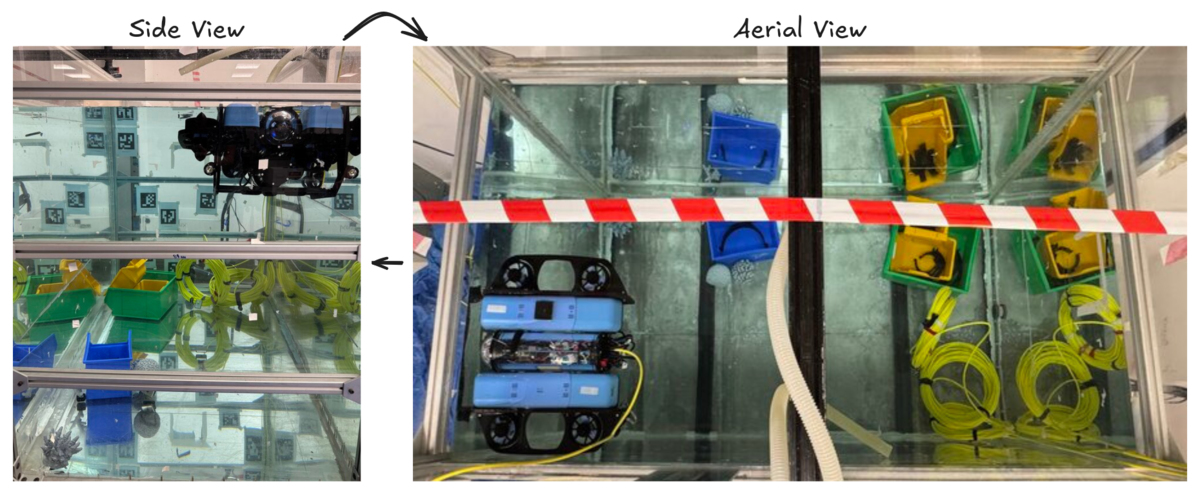}}
    \subcaptionbox{BlueROV2 (large pool) setup\label{fig:large_pool_setup}}{\includegraphics[height=1.65in]{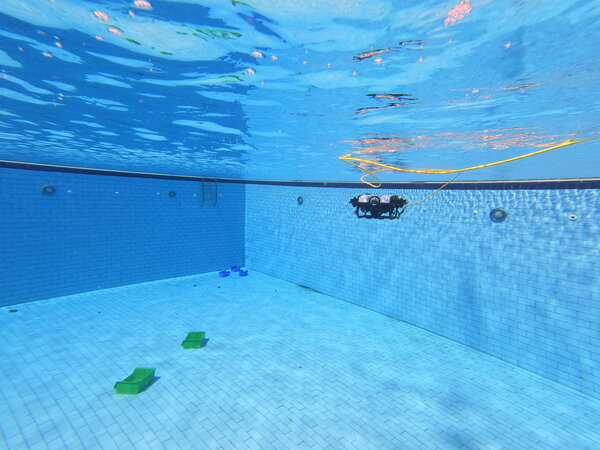}}
    \caption{Snapshots of the various setups experiments were conducted in. In all cases, the target was never visible from the starting position.}
    \label{fig:experiment_setup_pictures}
\end{figure*}

\subsection{Query-Driven Visual Perception}
\label{subsec:perception}

Our Monitor module reformulates perception as a language-conditioned visual question-answering process. At each step, the Controller first generates a task-relevant visual query $q_t$ based on its current context (task objective, action history, and environment memory). This query is dispatched to each Monitor $m_c$, which analyzes the corresponding camera image $I_t^{(c)}$ through a VLM and returns a natural language scene description. For low-resolution cameras (e.g., $96 \times 96$ pixels), a two-stage learned super-resolution pipeline upscales images before VLM inference.
% To mitigate VLM hallucination, each Monitor processes the same image $N$ times independently, producing $\{o_{t,1}^{(c)}, \dots, o_{t,N}^{(c)}\}$. A meta-analysis step then synthesizes these into a consolidated observation $o_t^{(c)}$ by favoring assertions consistent across runs and discarding isolated claims.

This design differs from conventional perception pipelines (object detectors, depth estimators) that produce fixed-schema numerical outputs (bounding boxes, class labels) regardless of task context. Three properties follow. First, perception is \textit{task-adaptive}: the query $q_t$ varies with the execution stage, so the Monitor attends to different scene aspects without architectural change. Second, the output resides in the \textit{same representational space} as the Controller's input, eliminating hand-designed encoders that would otherwise be embodiment-specific. Third, the VLM provides \textit{open-vocabulary} recognition without domain-specific training.

\subsection{LLM-Curated Persistent Memory}
\label{subsec:memory}

Naively appending all observations to the conversation history leads to unbounded context growth and eventual degradation. We address this with a dedicated Memory Curator, a separate LLM instance that maintains a structured, bounded representation of accumulated knowledge.

\paragraph{Incremental rewriting} After each control step, the Memory Curator receives the current memory $\mathcal{M}_{t-1}$ together with the latest interaction record (visual query, monitor responses, and executed action) and produces an updated memory $\mathcal{M}_t$. This is a \textit{rewriting} operation rather than appending: the Curator compresses redundant information, resolves contradictions between old and new observations, and discards details no longer relevant to the current task, keeping $\mathcal{M}_t$ bounded in length.

\paragraph{Structured schema} The Curator organizes knowledge into 4 categories: (1)~\emph{physical environment}, including scene description, spatial layout, and a dynamic object inventory with properties and last-observed locations; (2)~\emph{robot state}, covering position, orientation, and joint configuration; (3)~\emph{curated history} of significant commands and their outcomes, filtered for novelty and task relevance; and (4)~\emph{task state} with current objectives and completion status.

\paragraph{Cross-modal position inference} The memory system infers object positions by intersecting two cues: which camera observed the object and which action brought it into view. For example, if a rightward camera detects an object after a leftward movement, the object is inferred to be to the left of the robot's starting position. These relative spatial estimates are updated as the robot moves (e.g., an object recorded as ``in front'' becomes ``front-right'' after a leftward movement), providing spatial memory that compensates for the VLM's lack of metric depth estimation.

To validate the Memory Curator, we conducted a proof-of-concept experiment using the OpenAI Gym Blackjack environment~\cite{brockman2016openai}. Details are provided in the appendix.
% the Memory Curator capabilities were validated using the Gymnasium Blackjack environment, and the results are presented in the appendix.

\section{EXPERIMENTAL SETUP}\label{sec:experimental_setup}

% ~ 1 page

We evaluate the proposed framework across three platforms that span radically different morphologies, movement capabilities, and operating environments. Crucially, the control logic and LLM modules remain identical across all platforms; only the prompt configuration files (robot description $\mathcal{D}$, action definitions $\mathcal{A}$, and environment context) are varied. When it was safe to do so, a baseline is also established using randomly selected actions.

\subsection{Robot Platforms}
\label{sec:platforms}

\paragraph{Alhakami et al. Limb} A multi-jointed manipulator based on the work of \cite{alhakami2025towards} with  servo-driven axes: horizontal (left/right), vertical (up/down), rotational (clockwise/counter-clockwise), and a single-finger end effector (bend/extend). The arm is equipped with six onboard cameras (although we disabled the two side-facing cameras in our experiments to make the task more challenging) that operate at $100 \times 100$~px resolution. In order to achieve improved reasoning capabilities, the camera image has been up-sampled. The action space contains eight directional primitives plus a reset position command ($|\mathcal{A}|=9$). Low-resolution images are upscaled to $768 \times 768$~px via a two-stage Swin2SR super-resolution pipeline before VLM inference.

\paragraph{Clearpath Dingo} A wheeled ground robot evaluated in both NVIDIA Isaac Sim\footnote{\url{https://developer.nvidia.com/isaac/sim}} and in the physical environment shown in \cref{fig:experiment_setup_pictures}. In simulation, the robot is equipped with three cameras (forward, left, right) at $640 \times 480$~px. In the real-world environment, the robot is equipped with one forward-facing camera providing $1280 \times 800$~px image. In both cases, motion is restricted to the ground plane with actions: forward, backward, rotate left, and rotate right ($|\mathcal{A}|=4$). A proximity threshold of 1~m to the target object defines episode success.

\paragraph{BlueROV2 underwater vehicle} A thruster-driven remotely operated vehicle equipped with four vectored and four vertical thrusters, controlled through the MAVLink protocol using ArduSub firmware. A single forward-facing $1920 \times 1080$~px camera provides visual feedback over UDP using GStreamer. The vehicle is fully actuated in six degrees of freedom; however, for the purposes of this study, control is restricted to three translational/rotational axes (surge, heave, and yaw). The resulting discrete action set $\mathcal{A}$ consists of six commands: surge forward/backward, yaw left/right, and heave up/down ($|\mathcal{A}|=6$). Thrust magnitudes are configured per action (300--600 on a 0--1000 scale) with a 2-second command duration.

Table~\ref{tab:platforms} summarizes the key differences between platforms.

\begin{table}[t]
    \centering
    \vspace{4pt}
    \caption{Platform Configurations
    %. The same framework operates across all platforms with no code changes.
    }
    \resizebox{\columnwidth}{!}{%
    \begin{tabular}{|l|c|c|c|c|}
        \hline
        Platform & Morphology & \# Cam. & $|\mathcal{A}|$ & Img. res. \\
        \hline
        Alhakami et al. Limb & 4-DOF arm & 4 & 9 & $100 \times 100$ \\
        Dingo (simulation) & 3-DOF Wheeled & 3 & 4 & $640 \times 480$ \\
        Dingo (real-world) & 3-DOF Wheeled & 1 & 4 & $1280 \times 800$ \\
        BlueROV2 & 6-DOF ROV & 1 & 6 & $1920 \times 1080$ \\
        \hline
    \end{tabular}%
    }
    \label{tab:platforms}
\end{table}

\subsection{Tasks}
\label{sec:tasks}

\paragraph{Object localization (Alhakami et al. Limb)} The robot must locate a designated target object in a cluttered laboratory workspace by coordinating multi-camera observations and articulated joint movements. The object selected for the system to locate is a fire extinguisher as shown in \cref{fig:limb_setup}. The task is completed when a separate VLM detects it in any of the cameras' fields of view. 30 runs of the experiment were conducted.

\paragraph{Target approach (Dingo)} The robot navigates toward a fire extinguisher placed in a simulated $20 \times 30$~meters warehouse environment using only visual feedback. The robot camera views as well as the setup of the experiment are presented in \cref{fig:simulation_setup}. The episode terminates successfully when the robot reaches within 1~m of the target. A similar approach is implemented with the real-world version of the experiment. Robot operates in a $9.5 \times 5.5$~m area separated from the laboratory environment by net-like curtains as shown in \cref{fig:real_world_setup}. For both of the cases, the experiment was conducted 10 times.

\paragraph{Underwater navigation (BlueROV2)} The robot navigates to a designated target within a water tank measuring $2.00 \times 1.10 \times 1.50$~m, filled to $80\%$ of its volume. The operator specifies the target being a green box through a natural language instruction, and the robot must orient and maneuver in three dimensions using thruster commands and a single camera feed. The camera image is limited to $512 \times 512$~px taken from the top of the frame in order to limit the field of view. This alteration is applied in order to make the task harder by not allowing the floor of the tank to be visible from the surface level. The run was repeated 10 times.
The tank composition for the run can be seen in \cref{fig:small_tank_setup}. Subsequent experiments were conducted in a larger pool of size $ 15 \times 15 $~m with a depth of up to $3.8$~m. For this case, there is no image cropping present, and the target object is set to a group of blue boxes placed in the corner of the pool (see \cref{fig:large_pool_setup}). The pool-based experiments consisted of 5 runs.

\subsection{Prompt-Based Adaptation}
\label{sec:adaptation}

Adapting the framework to each platform requires authoring three text files: (1) a robot description $\mathcal{D}$ summarizing the morphology, joint structure, and sensor layout in natural language; (2) a tool definition file specifying the admissible actions $\mathcal{A}$ in a structured JSON schema; and (3) the task description. No source code, model weights, or reward functions are modified.
%For the three platforms evaluated, the robot descriptions range from 20 to 60 lines of text, and the tool definition files contain 2 to 10 action entries.

\subsection{Implementation Details}
\label{sec:implementation}

All LLM and VLM inference uses the OpenAI GPT-4.1-mini or GPT-4.1 models via the official API. The smaller model was applied to all experiments except Dingo (real-world) and BlueROV2 (large pool), where the larger version was used. The Controller and Memory Curator operate as text-only LLM calls; the Monitors operate as VLM calls with image input. The control loop executes at approximately one action per 5--10 seconds, dominated by API latency. The total cost of API calls amounted to less than \$100.

\section{RESULTS \& DISCUSSION}\label{sec:results_and_discussion}

% ~ 1.5 page

\begin{figure}[t]
    \centering
    \includegraphics[width=.8\linewidth]{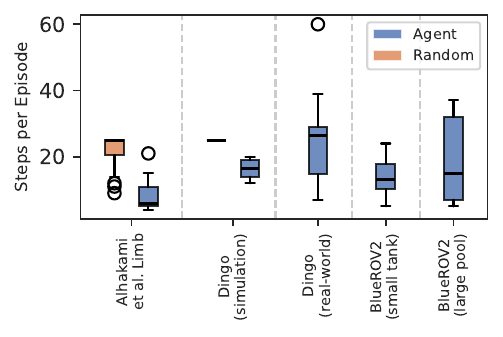}
    \caption{Distribution of steps needed to complete the tasks in each of the settings. Note that random control policies were artificially limited to 25 steps.}
    \label{fig:results}
\end{figure}

\cref{tab:results} shows the average number of steps needed to solve each setting compared with a random baseline (where relevant) and the minimal number of steps needed to solve the task.
Note that the minimum number of steps shown in \cref{tab:results} is under the assumption that an oracle is available, and so represents an unreachable upper-bound on performance.
In the large pool setting, additional stochasticity exists due to the presence of wind, leading to the optimal number of steps being approximate.
In cases where a random baseline was considered, the performance of RACAS was significantly better (family-wise confidence of $p < 0.01$ using Student's t-test and Holm-Bonferroni corrections).
\cref{fig:results} further shows the distribution of steps per episode in each of the settings.
Notably, despite the radically different environments and the added complexities, the settings where it was unsafe to run a random baseline were still generally able to solve the task in a reasonable number of steps.

\begin{table}[t]
    \centering
    \vspace{4pt}
    \caption{Performance Comparison of RACAS and Baselines ($\pm$ standard error)}
    \resizebox{\columnwidth}{!}{%
    \begin{tabular}{|l|c|c|c|}
        \hline
         Experiment & Baseline & RACAS & Min. \\
         \hline
         Dingo (simulation) & 25.00 $\pm$ 0.00 & 16.40 $\pm$ 0.96 & 10 \\
         Dingo (real-world) & N/A & 25.80 $\pm$ 4.94 & 7\\
         Alhakami et al. Limb & 22.07 $\pm$ 0.98 & 9.56 $\pm$ 1.50 & 3 \\
         BlueROV2 (small tank) & N/A & 14.00 $\pm$ 1.75 & 5\\
         BlueROV2 (large pool) & N/A & 19.20 $\pm$ 6.51 & $\approx$ 5\\
         \hline
    \end{tabular}%
    }
    \label{tab:results}
\end{table}

Altogether, these results demonstrate clear evidence that RACAS was able to cleanly solve all the tasks in a reasonable number of steps.
Thus, we can conclude that, at least for relatively simple navigation tasks, RACAS enables the generalization abilities of LLMs and VLMs to carry over to solving robotic tasks through relatively arbitrary morphologies and environments.

During operation, we noted that the time to complete the tasks was heavily dominated by the system attempting to locate any trace of the object.
Once spotted, RACAS consistently directed the robot in an almost straight line towards the objective.
This further suggests that the performance of RACAS is more limited by the sensor fidelity and available tools rather than by the underlying models or system architecture.

\section{CONCLUSION}\label{sec:conclusion}

We presented RACAS, a single agentic system that uses cooperative LLM and VLM modules to provide closed-loop control across heterogeneous robot embodiments.
All embodiment-specific knowledge is confined to declarative prompt configurations.
No source code, model weights, or reward functions need to be modified to move between platforms.
We evaluated RACAS on three platforms that differ radically in morphology, maneuverability, and operating environment, including a recently-published robotic limb that an LLM would have no prior knowledge of.
In all cases, the unmodified system completed the assigned tasks.
% To the best of our knowledge, this is the first such demonstration.
To the best of our knowledge, this constitutes the first zero-training generalization of a unified control framework across such fundamentally different robotic platforms.

% The conventional approach to deploying a robot on a new task requires collecting demonstration data, engineering state representations, and training platform-dependent policies. This is expensive, and the expertise needed spans disjoint specializations.
% RACAS sidesteps this entirely for the target finding tasks studied here: a practitioner needs only a robot's API and the ability to write natural language descriptions.
% Altogether, this points to a meaningful reduction in the barrier to prototyping robotic solutions, one that will only grow as the underlying foundation models improve.

RACAS sidesteps the traditional barriers to robotic deployment that arise from collecting demonstration data, engineering custom state representations, and training platform-specific policies. Instead, a practitioner can adapt the system to a new robot using only the robot's existing API and natural language descriptions. This represents a significant reduction in the expertise and resources needed to prototype robotic solutions, and as foundation models continue to improve in their reasoning and perception capabilities, we expect this approach to become increasingly viable for a wider range of tasks and platforms.

\section{LIMITATIONS AND FUTURE WORK}\label{sec:limitations_and_future_work}
% The key limitation of this work is that RACAS is only implemented on target localization and navigation tasks.
% While sufficient for the purposes of evaluating the potential of RACAS to enable the generalization of LLMs and VLMs to carry over, this only covers one part of a usual robotic task.
% However, tasks such as manipulation are notoriously complicated, making experiments with high inference costs at each step prohibitively slow.
% Future work will focus on integrating existing manipulator systems as specific tools for the agentic system to leverage.

Our evaluation is restricted to target localization and navigation tasks, which, while sufficient for demonstrating cross-embodiment generalization, represent only one stage of a typical robotic mission. Extending RACAS to contact-rich tasks such as manipulation is a natural next step, though the high per-step inference cost of the current architecture makes long-horizon manipulation experiments prohibitively slow. Future work will explore the integration of existing manipulation primitives as callable tools within the agentic framework, reducing the number of LLM-mediated decisions per task.

A second limitation, one consistent issue that harmed the abilities of the system was a lack of depth information. While this was able to be inferred from the images, such inferences were sub-optimal, leading to the robots being under- or over-confident on impact chances. Future work will look at integrating either LIDAR or a more sophisticated architecture into the monitor sub-systems.


\appendix

\section{Memory Curator Test}\label{app:memory_curator}

To evaluate the Memory Curator's ability to accumulate and leverage experience across episodes, the system is applied to a modified version of the Blackjack environment from OpenAI Gym~\cite{brockman2016openai}. Blackjack is well-suited for this purpose: its rules are simple enough that requiring no complex multi-agent coordination, yet optimal play hinges on an effective memory mechanism to infer a hidden numerical target from repeated outcomes. In the game, a player draws cards and must decide whether to take another card (hit) or stop (stick), aiming to reach the target sum without exceeding it.
The standard target of 21 is well-represented in LLM training corpora, meaning an LLM-based agent could play near-optimally from prior knowledge alone without relying on the Memory Curator. We therefore replace the target with 42, a value the agent is not informed of.  The agent is only told that the target lies in the range $X \in [12, 100]$ and must infer it from game outcomes across episodes.
% The action space consists of two actions: hit and stick ($|\mathcal{A}|=2$).

\begin{figure}[h]
    \centering
    \subcaptionbox{Camera view}{\includegraphics[width=0.25\linewidth]{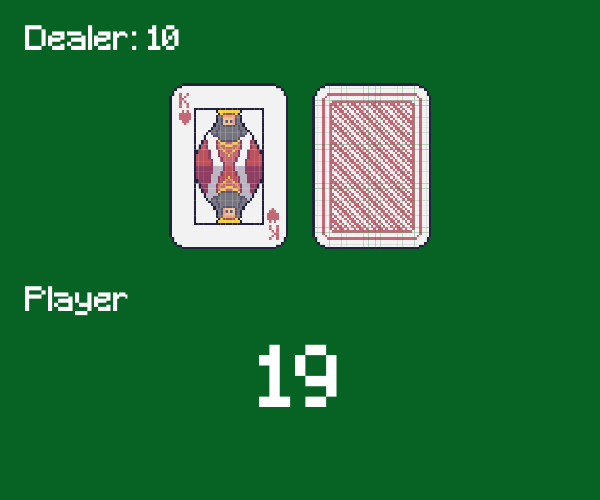}}
    \subcaptionbox{Cumulative average scores}{\includegraphics[width=0.65\linewidth,trim=0 0.75em 0 0,clip]{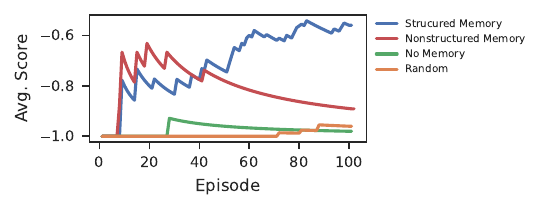}}
    \caption{Blackjack visualization and experimental results.}
    \label{fig:blackjack}
\end{figure}

In the game, the Monitor was given the rendered game-state images (\cref{fig:blackjack}) to identify card values, while the Controller reasons about optimal play strategies.

To provide a baseline for the experiment, a series of tests is run. Firstly, the Memory Curator module is disabled: the Controller receives the environment memory $\mathcal{M}_{t-1}$ in a non-structured manner, appending its previous outputs back with the input query. The second baseline is obtained through complete memory removal. The system must take actions based solely on current visual observations. This ablation measures whether persistent structured memory improves decision quality over reactive control. All tests were compared to a random choice policy.

The result shown in \cref{fig:blackjack} confirms the ability of the system to infer feedback from previous episodes of operation based on memory updates. Memory Curator allows the system to keep performing well with multiple instances of memory updates, allowing it to maintain high performance.

\section*{ACKNOWLEDGMENT}
The authors would like to thank David Alvear Goyes for his assistance in setting up the Dingo experiments.
This work was supported by the Center of Excellence for Generative AI at the King Abdullah University of Science and Technology (KAUST, Award Number 5940), the European Research Council (ERC, Advanced Grant Number 742870), and the Swiss National Science Foundation (SNF, Grant Number 200021 192356).
Large language models were used throughout the paper to help draft sections and improve readability.

%%%%%%%%%%%%%%%%%%%%%%%%%%%%%%%%%%%%%%%%%%%%%%%%%%%%%%%%%%%%%%%%%%%%%%%%%%%%%%%%

\bibliographystyle{IEEEtran}
\bibliography{dylan,main}

\begin{thebibliography}{10}
\providecommand{\url}[1]{#1}
\csname url@rmstyle\endcsname
\providecommand{\newblock}{\relax}
\providecommand{\bibinfo}[2]{#2}
\providecommand\BIBentrySTDinterwordspacing{\spaceskip=0pt\relax}
\providecommand\BIBentryALTinterwordstretchfactor{4}
\providecommand\BIBentryALTinterwordspacing{\spaceskip=\fontdimen2\font plus
\BIBentryALTinterwordstretchfactor\fontdimen3\font minus \fontdimen4\font\relax}
\providecommand\BIBforeignlanguage[2]{{%
\expandafter\ifx\csname l@#1\endcsname\relax
\typeout{** WARNING: IEEEtran.bst: No hyphenation pattern has been}%
\typeout{** loaded for the language `#1'. Using the pattern for}%
\typeout{** the default language instead.}%
\else
\language=\csname l@#1\endcsname
\fi
#2}}

\bibitem{brohan2022rt1}
A.~Brohan, N.~Brown, J.~Carbajal, Y.~Chebotar, J.~Dabis, C.~Finn, K.~Gopalakrishnan, K.~Hausman, A.~Herzog, \emph{et~al.}, ``Rt-1: Robotics transformer for real-world control at scale,'' \emph{arXiv preprint arXiv:2212.06817}, 2022.

\bibitem{brohan2023rt2}
------, ``Rt-2: Vision-language-action models transfer web knowledge to robotic control,'' in \emph{7th Annual Conference on Robot Learning (CoRL)}, 2023.

\bibitem{bousmalis2023robocat}
K.~Bousmalis, G.~Vezzani, D.~Rao, C.~Devin, A.~X. Lee, M.~Bauza, T.~Davchev, Y.~Zhou, A.~Gupta, \emph{et~al.}, ``Robocat: A self-improving generalist agent for robotic manipulation,'' \emph{arXiv preprint arXiv:2306.11706}, 2023.

\bibitem{vemprala2024chatgptforrobotics}
S.~H. Vemprala, R.~Bonatti, A.~Bucker, and A.~Kapoor, ``Chatgpt for robotics: Design principles and model abilities,'' \emph{IEEE Access}, vol.~12, pp. 55\,682--55\,696, 2024.

\bibitem{ahn2022saycan}
M.~Ahn, A.~Brohan, N.~Brown, Y.~Chebotar, O.~Cortes, B.~David, C.~Finn, C.~Fu, K.~Gopalakrishnan, \emph{et~al.}, ``Do as i can, not as i say: Grounding language in robotic affordances,'' in \emph{6th Annual Conference on Robot Learning (CoRL)}, 2022, pp. 287--318.

\bibitem{liang2023codeaspolicies}
J.~Liang, W.~Huang, F.~Xia, P.~Xu, K.~Hausman, B.~Ichter, P.~Florence, and A.~Zeng, ``Code as policies: Language model programs for embodied control,'' in \emph{2023 IEEE International Conference on Robotics and Automation (ICRA)}, 2023, pp. 9493--9500.

\bibitem{schmidhuber2015learning}
J.~Schmidhuber, \emph{On Learning to Think: Algorithmic Information Theory for Novel Combinations of Reinforcement Learning Controllers and Recurrent Neural World Models}.\hskip 1em plus 0.5em minus 0.4em\relax ar{X}iv, 2015.

\bibitem{zhuge2023mindstorms}
M.~Zhuge, H.~Liu, F.~Faccio, D.~R. Ashley, R.~Csord{\'{a}}s, A.~Gopalakrishnan, A.~Hamdi, H.~A. A.~K. Hammoud, V.~Herrmann, \emph{et~al.}, \emph{Mindstorms in Natural Language-Based Societies of Mind}.\hskip 1em plus 0.5em minus 0.4em\relax ar{X}iv, 2023.

\bibitem{zhuge2025mindstorms}
------, ``Mindstorms in natural language-based societies of mind,'' \emph{Computational Visual Media}, vol.~11, no.~1, pp. 29--81, 2025.

\bibitem{alhakami2025towards}
M.~Alhakami, D.~R. Ashley, J.~Dunham, Y.~Dai, F.~Faccio, E.~Feron, and J.~Schmidhuber, ``Towards an extremely robust baby robot with rich interaction ability for advanced machine learning algorithms,'' \emph{Proceedings of the 2025 {IEEE/RSJ} International Conference on Intelligent Robots and Systems}, 2025.

\bibitem{huang2022innermonologue}
W.~Huang, F.~Xia, T.~Xiao, H.~Chan, J.~Liang, P.~Florence, A.~Zeng, J.~Tompson, I.~Mordatch, \emph{et~al.}, ``Inner monologue: Embodied reasoning through planning with language models,'' in \emph{6th Annual Conference on Robot Learning (CoRL)}, 2022.

\bibitem{raptis2025robotiq}
E.~K. Raptis, A.~C. Kapoutsis, and E.~B. Kosmatopoulos, ``Robotiq: Empowering mobile robots with human-level planning for real-world execution,'' \emph{arXiv preprint arXiv:2502.12862}, 2025.

\bibitem{singh2024malmm}
H.~Singh, R.~J. Das, M.~Han, P.~Nakov, and I.~Laptev, ``Malmm: Multi-agent large language models for zero-shot robotic manipulation,'' in \emph{2025 IEEE/RSJ International Conference on Intelligent Robots and Systems (IROS)}, 2025, pp. 20\,386--20\,393.

\bibitem{mandi2024roco}
Z.~Mandi, S.~Jain, and S.~Song, ``Roco: Dialectic multi-robot collaboration with large language models,'' in \emph{2024 IEEE International Conference on Robotics and Automation (ICRA)}, 2024, pp. 286--299.

\bibitem{strobel2024llm2swarm}
V.~Strobel, M.~Dorigo, and M.~Fritz, ``Llm2swarm: Robot swarms that responsively reason, plan, and collaborate through llms,'' in \emph{NeurIPS 2024 Workshop on Open-World Agents (OWA-2024)}, 2024.

\bibitem{duan2024manipulateanything}
J.~Duan, W.~Yuan, W.~Pumacay, Y.~R. Wang, K.~Ehsani, D.~Fox, and R.~Krishna, ``Manipulate-anything: Automating real-world robots using vision-language models,'' in \emph{8th Annual Conference on Robot Learning (CoRL)}, 2024.

\bibitem{ahn2024vader}
M.~Ahn, M.~Gonzalez~Arenas, M.~Bennice, N.~Brown, C.~Chan, B.~David, A.~Francis, G.~Gonzalez, R.~Hessmer, \emph{et~al.}, ``Vader: Visual affordance detection and error recovery for multi robot human collaboration,'' \emph{arXiv preprint arXiv:2405.16021}, 2024.

\bibitem{finn2017maml}
C.~Finn, P.~Abbeel, and S.~Levine, ``Model-agnostic meta-learning for fast adaptation of deep networks,'' in \emph{Proceedings of the 34th International Conference on Machine Learning (ICML)}, 2017.

\bibitem{yu2020metaworld}
T.~Yu, D.~Quillen, Z.~He, R.~Julian, K.~Hausman, C.~Finn, and S.~Levine, ``Meta-world: A benchmark and evaluation for multi-task and meta reinforcement learning,'' in \emph{Proceedings of the 3rd Annual Conference on Robot Learning (CoRL)}, 2020.

\bibitem{beck2025tutorial}
J.~Beck, R.~Vuorio, E.~Zheran~Liu, Z.~Xiong, L.~Zintgraf, C.~Finn, and S.~Whiteson, ``A tutorial on meta-reinforcement learning,'' \emph{Foundations and Trends in Machine Learning}, vol.~18, no. 2-3, pp. 224--384, 2025.

\bibitem{firoozi2025foundation}
R.~Firoozi, J.~Tucker, S.~Tian, A.~Majumdar, J.~Sun, W.~Liu, Y.~Zhu, S.~Song, A.~Kapoor, \emph{et~al.}, ``Foundation models in robotics: Applications, challenges, and the future,'' \emph{The International Journal of Robotics Research}, vol.~44, no.~5, pp. 701--739, 2025.

\bibitem{driess2023palme}
D.~Driess, F.~Xia, M.~S.~M. Sajjadi, C.~Lynch, A.~Chowdhery, B.~Ichter, A.~Wahid, J.~Tompson, Q.~Vuong, \emph{et~al.}, ``Palm-e: An embodied multimodal language model,'' in \emph{International Conference on Machine Learning (ICML)}, 2023, pp. 8469--8488.

\bibitem{reed2022gatox}
S.~Reed, K.~Zolna, E.~Parisotto, S.~Gomez~Colmenarejo, A.~Novikov, G.~Barth-Maron, M.~Gimenez, Y.~Sulsky, J.~Kay, \emph{et~al.}, ``A generalist agent,'' \emph{arXiv preprint arXiv:2205.06175}, 2022.

\bibitem{rana2023sayplan}
K.~Rana, J.~Haviland, S.~Garg, J.~Abou-Chakra, I.~Reid, and N.~S{\"u}nderhauf, ``Sayplan: Grounding large language models using 3d scene graphs for scalable task planning,'' in \emph{7th Annual Conference on Robot Learning}, 2023.

\bibitem{ding2023tampobjectrearrangement}
Y.~Ding, X.~Zhang, C.~Paxton, and S.~Zhang, ``Task and motion planning with large language models for object rearrangement,'' in \emph{2023 IEEE/RSJ International Conference on Intelligent Robots and Systems (IROS)}, 2023, pp. 2086--2092.

\bibitem{singh2023progprompt}
I.~Singh, V.~Blukis, A.~Mousavian, A.~Goyal, D.~Xu, J.~Tremblay, \emph{et~al.}, ``Progprompt: Generating situated robot task plans using large language models,'' in \emph{2023 IEEE International Conference on Robotics and Automation (ICRA)}, 2023, pp. 11\,523--11\,530.

\bibitem{huang2023voxposer}
W.~Huang, C.~Wang, R.~Zhang, Y.~Li, J.~Wu, and L.~Fei-Fei, ``Voxposer: Composable 3d value maps for robotic manipulation with language models,'' in \emph{7th Annual Conference on Robot Learning (CoRL)}, 2023.

\bibitem{wang2024ronar}
Z.~Wang, B.~Liang, V.~Dhat, Z.~Brumbaugh, N.~Walker, R.~Krishna, and M.~Cakmak, ``I can tell what i am doing: Toward real-world natural language grounding of robot experiences,'' in \emph{8th Annual Conference on Robot Learning (CoRL)}, 2024.

\bibitem{stone2023moo}
A.~Stone, T.~Xiao, Y.~Lu, K.~Gopalakrishnan, K.-H. Lee, Q.~Vuong, P.~Wohlhart, S.~Kirmani, B.~Zitkovich, \emph{et~al.}, ``Open-world object manipulation using pre-trained vision-language models,'' in \emph{7th Annual Conference on Robot Learning (CoRL)}, 2023, pp. 3397--3417.

\bibitem{shah2024bumble}
R.~Shah, A.~Yu, Y.~Zhu, Y.~Zhu, and R.~Mart{\'\i}n-Mart{\'\i}n, ``Bumble: Unifying reasoning and acting with vision-language models for building-wide mobile manipulation,'' \emph{arXiv preprint arXiv:2410.06237}, 2024.

\bibitem{qu2026pragmabot}
K.~Qu, G.~Lan, R.~Zurbr{\"u}gg, C.~Chen, C.~E. Mower, H.~Bou-Ammar, and M.~Hutter, ``A pragmatist robot: Learning to plan tasks by experiencing the real world,'' \emph{arXiv preprint arXiv:2507.16713}, 2026.

\bibitem{brockman2016openai}
G.~Brockman, V.~Cheung, L.~Pettersson, J.~Schneider, J.~Schulman, J.~Tang, and W.~Zaremba, \emph{{OpenAI} Gym}.\hskip 1em plus 0.5em minus 0.4em\relax ar{X}iv, 2016.

\end{thebibliography}

\end{document}